%% file: root.tex
\documentclass[10pt,a4paper,conference]{IEEEtran}
\usepackage[utf8]{inputenc}

\usepackage{hyperref}
\usepackage{csquotes}
\usepackage{todonotes}
\usepackage{xfrac}  

\usepackage{tikz}
\usetikzlibrary{calc}
\usetikzlibrary{arrows.meta}

\usepackage{cite}
\usepackage{amsmath,amssymb,amsfonts}
\usepackage{isomath}
\ifCLASSOPTIONcompsoc
  \usepackage[caption=false,font=normalsize,labelfont=sf,textfont=sf]{subfig}
\else
  \usepackage[caption=false,font=footnotesize]{subfig}
\fi

\usepackage{stfloats}
\renewcommand{\vec}{\mathbf}

\newcommand{\set}{\mathcal}

\newcommand{\stimes}{{\mkern-2mu\times\mkern-2mu}}

\definecolor{c1}{HTML}{1E91D6}
\definecolor{c2}{HTML}{0072BB}
\definecolor{c3}{HTML}{8FC93A}
\definecolor{c4}{HTML}{E4CC37}
\definecolor{c5}{HTML}{E18335}

\definecolor{b1}{HTML}{4DA3D6}
\definecolor{b2}{HTML}{2E84BA}
\definecolor{b3}{HTML}{9EC95E}
\definecolor{b4}{HTML}{E2D161}
\definecolor{b5}{HTML}{E09A60}

\begin{document}
%
\title{Image-based OoD-Detector Principles on Graph- based Input Data in Human Action Recognition}

\author{\IEEEauthorblockN{Jens Bayer~\href{https://orcid.org/0000-0002-2806-6920}{\includegraphics[scale=0.5]{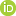}}}
\IEEEauthorblockA{
Fraunhofer Center for Machine\\ Learning and
Fraunhofer IOSB\\
Ettlingen, Germany\\
jens.bayer@iosb.fraunhofer.de
}
\and
\IEEEauthorblockN{David Münch~\href{https://orcid.org/0000-0002-8577-5256}{\includegraphics[scale=0.5]{images/orcid.png}}}
\IEEEauthorblockA{
Fraunhofer Center for Machine\\ Learning and
Fraunhofer IOSB\\
Ettlingen, Germany\\
david.muench@iosb.fraunhofer.de
}
\and
\IEEEauthorblockN{Michael Arens~\href{https://orcid.org/0000-0002-7857-0332}{\includegraphics[scale=0.5]{images/orcid.png}}}
\IEEEauthorblockA{
Fraunhofer Center for Machine\\ Learning and
Fraunhofer IOSB\\
Ettlingen, Germany\\
michael.arens@iosb.fraunhofer.de
}
}


%


\maketitle

\begin{abstract}
Living in a complex world like ours makes it unacceptable that a practical implementation of a machine learning system assumes a closed world.
Therefore, it is necessary for such a learning-based system in a real world environment, to be aware of its own capabilities and limits and to be able to distinguish between confident and unconfident results of the inference, especially if the sample cannot be explained by the underlying distribution.
This knowledge is particularly essential in safety-critical environments and tasks e.g. self-driving cars or medical applications.
Towards this end, we transfer image-based Out-of-Distribution (OoD)-methods to graph-based data and show the applicability in action recognition.

The contribution of this work is (i) the examination of the portability of recent image-based OoD-detectors for graph-based input data, (ii) a Metric Learning-based approach to detect OoD-samples, and (iii) the introduction of a novel semi-synthetic action recognition dataset.

The evaluation shows that image-based OoD-methods can be applied to graph-based data.
Additionally, there is a gap between the performance on intraclass and intradataset results. First methods as the examined baseline or ODIN provide reasonable results. More sophisticated network architectures -- in contrast to their image-based application -- were surpassed in the intradataset comparison and even lead to less classification accuracy.
\end{abstract}


%
\IEEEpeerreviewmaketitle

\section{Introduction}
Modern deep convolutional neural networks are able to recognize objects in images, segment areas pixelwise, and even generate realistic looking photos. Despite their superb capabilities in those areas, they are not able to exposure their own lack of knowledge. As some studies have found out, the confidence of a a network in its output is as high for irrelevant or non-human understandable input data as for in-distribution input data~\cite{Szegedy2013, Hendrycks2016,Nguyen2014}. As a result, there are numerous different approaches~\cite{Hendrycks2016, Liang2017, DeVries2018, Kliger2018, Masana2018} detecting so called out-of-distribution (OoD) data.

Instead of proposing another image based approach, this work investigates the applicability of OoD-detection methods on graph-based input data. 
To the best of our knowledge there are no OoD-detection methods which are usable and have been investigated on graph-based data.
Since human skeleton graphs can be easily generated from RGB images~\cite{Cao2018,Papandreou2018}, depth data~\cite{Papadopoulos2014}, and even RF-signals~\cite{Zhao2018a}, the representation of the dynamics of human actions can be captured without the high computational cost of optical flow or problems regarding poor visual conditions.
The contribution of this work is: (i) the examination of the portability of ODIN~\cite{Liang2017} and the confidence learning approach from \cite{DeVries2018}, when using graph-structured input data in an action recognition task. As a baseline, the softmax output comparison proposed in \cite{Hendrycks2016} is used. Additionally, (ii) a Metric Learning-based approach detecting OoD-samples is developed. (iii) To ensure to have a controlled and repeatable evaluation environment, a novel semi-synthetic action recognition dataset is also introduced.

In the following section an overview of related work on both graph-based structured action recognition and OoD-detection is given. The baseline method and the examined methods are explained in \autoref{sec:ood}. The semi-synthetic dataset and the quantitative evaluation are presented in \autoref{sec:eval}.


\begin{figure}[htb]
    \centering
    \includegraphics[width=\linewidth]{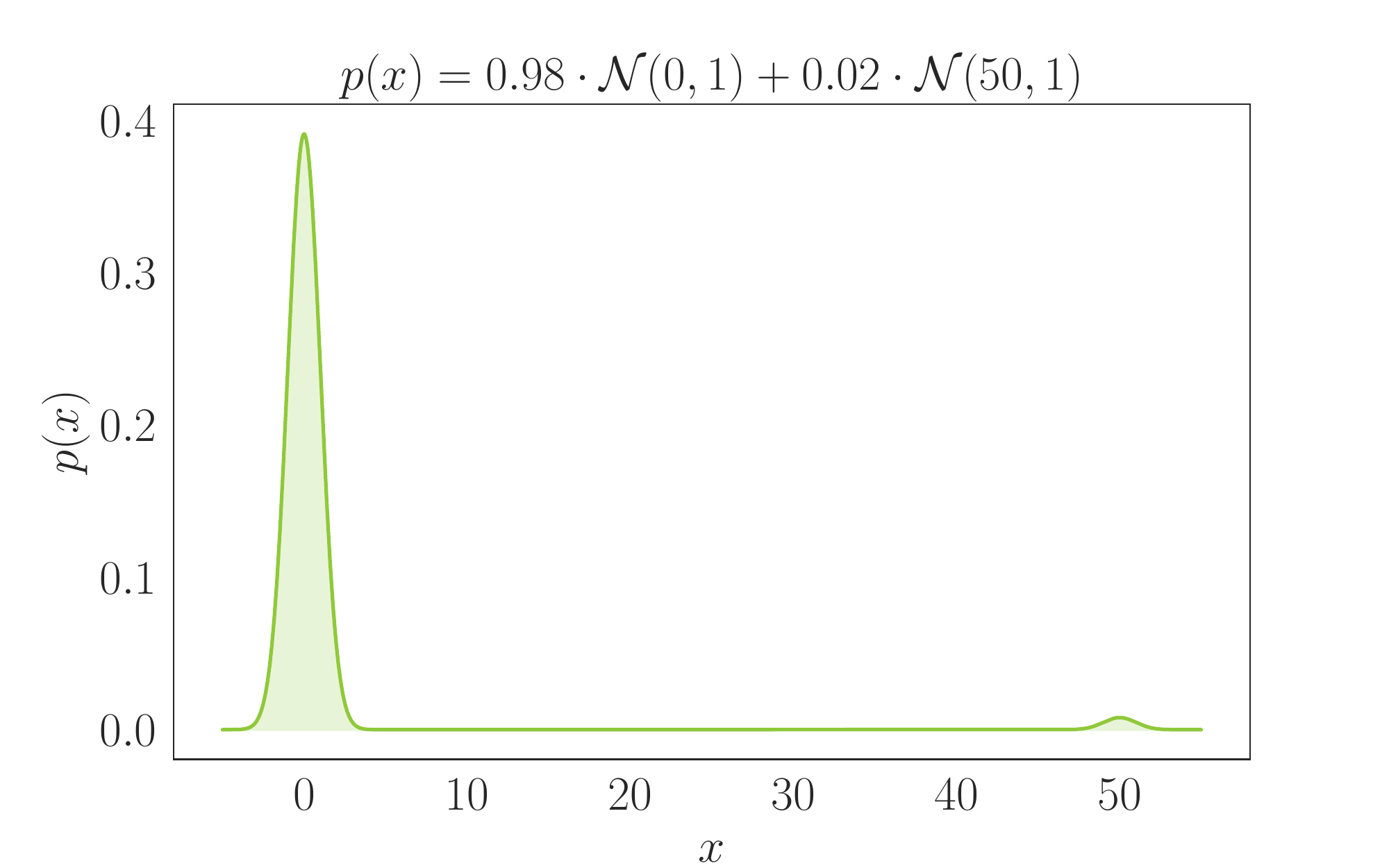}
    \caption{The definition of OoD-data is mandatory. Values in the range $x=[-47,53]$ are explainable by the given distribution but significantly less common then values in the range $x=[-3,3]$.}
    \label{fig:inlierproblem}
\end{figure}

\section{Related Work}
\label{sec:relatedwork}
Both in action recognition and outlier detection there is a large number of related work. We focus on skeleton-based action recognition as well as deep neural network outlier detection approaches.
However, additional information regarding action recognition can be found in the surveys~\cite{Turaga2008, Poppe2010, Kong2018}. A good overview on outlier detection is given by~\cite{Hodge2004, Bhosale2014, Zimek2018}.

\subsection{Skeleton-based Action Recognition}
\begin{figure}[tb]
    \centering
    \subfloat[]{
        \includegraphics[height=5.5cm]{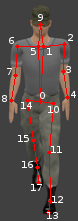}
    }
    \qquad
    \subfloat[]{
        \includegraphics[height=5.5cm]{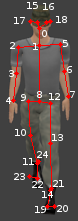}
    }
    \caption{Available skeleton data in the dataset: 18 node ground truth data (a) and 25 node OpenPose generated data (b).}
    \label{fig:datasetgraphbasedata}
\end{figure}

Recognizing actions based on image data is one way to solve action recognition tasks. Another strategy uses skeleton data which can be extracted by a 2D or 3D pose estimator such as Stacked Hourglass Networks~\cite{Newell2016}, PersonLab~\cite{Papandreou2018}, or OpenPose~\cite{Cao2018}.
The extracted landmarks can be seen as human joints and form the nodes of a skeleton graph (\autoref{fig:datasetgraphbasedata}). Based upon a time series of this graph input data, there are several ways on how to recognize an action.

A common approach is the analysis and classification of hand-crafted features using Hidden Markov Models~\cite{Papadopoulos2014}, Support Vector Machines~\cite{Kerola2015}, or k-Nearest-Neighbor classifiers~\cite{Devanne2015}.
Deep learning models~\cite{YongDu2015, Shahroudy2016, Yan2018, Si2018} are trained in an end-to-end manner and do not rely on handcrafted features.

In \cite{YongDu2015}, the skeleton graph is divided into five parts according to the human physical structure. These five parts are then fed separately into five bidirectional recurrent subnets (BRNN). The outputs of the subnets are successive fused to be the input of higher BRNN layers and build finally the input of the classification layer.

Part-aware LSTM networks are introduced in~\cite{Shahroudy2016}. A part-aware LSTM splits the entire motion of the human body into multiple part-based LSTM cells. To keep the context of each body part separated from one another, each cell has its individual input, forget, and modulation gates. Only the output gate is shared among all body parts. 

An approach using spatial temporal graph convolutional networks (ST-GCN) is given by \cite{Yan2018}. The skeleton sequence is interpreted as a graph in such a way, that in each frame, the corresponding joints of naturally connected joints in a human body are connected by an edge. To include the temporal domain, the same joints between consecutive frames share an edge.
The resulting graph is then propagated through the proposed graph convolution network which forms the input of the final classification layer.

Spatial reasoning and temporal stack learning networks are used in~\cite{Si2018}. While the former models high-level spatial structural information within each frame, the latter is responsible for generating detailed temporal dynamics.
A spatial reasoning network encodes the coordinate vector of each body part and feeds them into a residual graph neural network, which models the structural relationship between body parts. Those relationships are then analyzed in the temporal stack learning network, which stacks previous high-level features to generate even more high-level features. Based on the most high-level features, the system classifies an action.

\subsection{OoD-Detection}
Since detecting OoD-samples is an established topic, there are numerous detection methods which~\cite{Hodge2004} categorizes into statistical~\cite{Grubbs1969, Knorr1998, laurikkala2000informal, Allan1998, Seheult1989, DipankarDasgupta1996, Tax1999}, machine learning~\cite{John1995, Ester1996}, and neural network~\cite{Bishop1994, Nathalie1995, Hendrycks2016, Liang2017, DeVries2018, Kliger2018, Masana2018} based methods.

Assuming normal distributed data, \cite{Grubbs1969} calculates the mean and standard deviation of an attribute among all given data. An outlier is present, if the difference of the querying data and the mean divided by the standard deviation is lower than a predefined significance level.

To be able to handle multivariate data, \cite{laurikkala2000informal} uses the Mahalanobis distance to handle possible inter-attribute dependencies. The outlier detection is then performed by generating a boxplot based upon the calculated distance.

A biologically inspired method to detect novelty is presented by \cite{DipankarDasgupta1996} and uses an ensemble of simple detectors. Each detector checks the given data against its own definition of normality. If a detector detects an abnormal state, novelty has been detected.

To detect inlier,~\cite{Tax1999} uses a Support Vector Machine where the decision boundary is given by the sphere with minimal volume containing all data.

A method based on decision trees is presented by~\cite{John1995}, where a decision tree is repeatedly constructed and pruned. After each pruning step, incorrect classified samples are removed from the training set and are marked as outliers.

Some early neural network based methods are given by \cite{Bishop1994} and \cite{Nathalie1995}. The former takes advantage of the fact that a multilayer perceptron (MLP) works well for interpolating but bad for extrapolating data. More precisely, the MLP models the unconditional probability density of the input data used during training~\cite{Bishop1994}. The latter trains an autoencoder based on the training data. If the system is not able to sufficiently reconstruct the input during the test phase, the input is marked as OoD~\cite{Nathalie1995}.

A more recent neural network method is given by~\cite{Hendrycks2016} who propose to check the maximum value of the softmax output of a classifying neural network against a predefined threshold. If the maximum is below the threshold, the system marks the input as an outlier. The authors mention that this method can be considered as a baseline, as it is the most naïve way to decide whether an in- or an outlier is present.

The method proposed in~\cite{Liang2017} can be seen as an extension of the baseline method above. It only differs in the use of the tempered softmax~\cite{Hinton2015} during the test phase. Since the network is trained with the default softmax, the tempered softmax (parameterized with high temperatures) forces the network to be sure with its decisions during the test phase.

In~\cite{DeVries2018}, a confidence estimation branch is appended to the network. This branch enables the network to directly estimate a degree of confidence instead of just classify the input in- or out-of-distribution. During the training, an additional confidence loss is added to prevent the network from being doubtful. The trained network is then able to provide an additional confidence output for a given input.

Another method modifying the basic network is given by~\cite{Masana2018}. Instead of a confidence branch, the presented extension maps the basic output onto a manifold and enables the possibility of using the Euclidean distance as a measure of out-of-distributioness.

\section{Out-of-Distribution Detectors}
\label{sec:ood}
Our proposed method is inspired by the metric learning~\cite{Masana2018} and confidence learning~\cite{DeVries2018} methods and tries to enable the network to estimate the local density around a sample in the estimated manifold. 

We start with a definition of OoD-samples:
A naïve definition is that OoD-samples are not explainable by an underlying learned distribution.
As shown in \autoref{fig:inlierproblem}, this interpretation is problematic. Even if the values between -47 and 53 are explainable by the given distribution, the likelihood of having a sample in this range is negligibly small.
Therefore, this work requires an in-distribution sample to be significantly explainable by the learned distribution.

For OoD-samples, \cite{Masana2018} distinguishes between novelty and anomaly based OoD-samples. While the former describes samples sharing some common space with the trained distribution, the latter includes samples that are not related with the trained distribution. Credit card fraud, terrorist activities, and system failures are prominent examples of anomalies of high interest~\cite{Chandola2009}.
A third category ignored by~\cite{Masana2018} are plain outliers that are neither part of a new class nor part of an anomaly. They simply lie on or beyond the decision borders for their classes. This can be the result of bad training data or insufficient training.

The experimental setup of this work can be seen as novelty detection problem: A predetermined single class is excluded during the training and only present during the test phase. The predetermined class can be seen as the OoD-class and should be rejected by the system.

\subsection{Baseline}
The approach presented in~\cite{Hendrycks2016} is used as a baseline. Given a pre-trained classifier which uses a softmax output layer, the proposed OoD-detector simply checks the maximum softmax output against a predefined threshold. If the maximum is greater than the threshold, the system continues its classification task. Otherwise the input is marked as OoD and hence rejected.
The threshold value is determined in such a way, that an error of $5\%$ is allowed. Therefore, the true positive rate is fixed at $95\%$. Except for the threshold determination, this method is one of the most naïve ways on handling the detection of OoD-samples.

\subsection{Out-of-DIstribution detector for Neural networks}
\begin{figure}
    \centering
    \includegraphics[width=\linewidth]{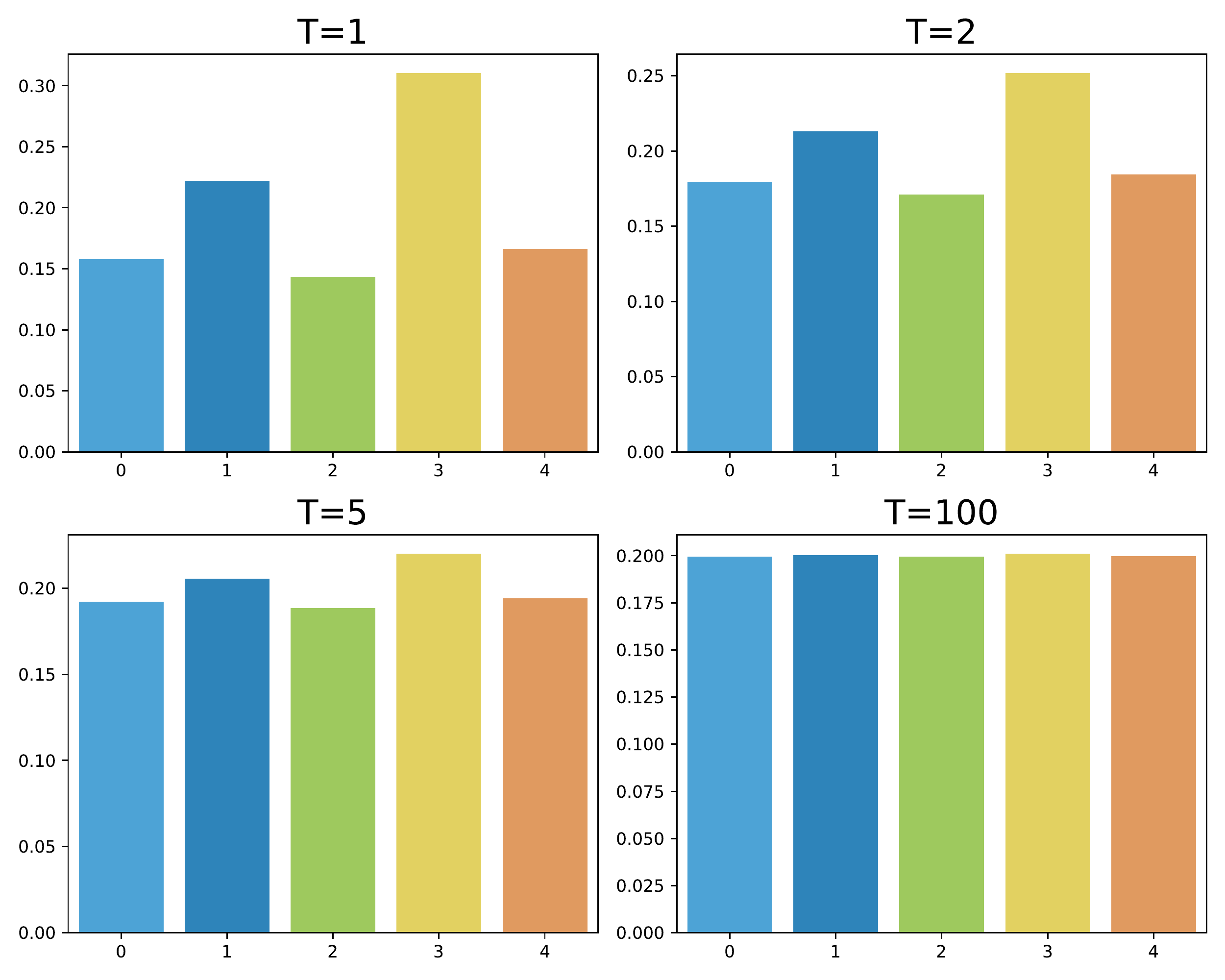}
    \caption{Tempered softmax applied to the same input with different values for $T$. The higher the temperature parameter, the more equally distributed is the output. Note range of the y-axis.}
    \label{fig:tmpsoftmax}
\end{figure}

ODIN~\cite{Liang2017} can be seen as an extension of the baseline approach. Inspired by \cite{Hinton2015}, the tempered softmax
\begin{align}
    \sigma_T (x)_c = \frac{e(\sfrac{x_c}{T})}{\sum_{j \in \set{C}}e(\sfrac{x_j}{T})}, \quad c \in \set{C}
    \label{eq:tmpsoftmax}
\end{align}
\vspace{-2pt}
is used during the test phase. The higher the temperature parameter $T$, the more equally distributed is its output among all available classes $\set{C}$ (see \autoref{fig:tmpsoftmax}). As a result, a high temperature parameter during the test phase forces the network to be confident in its classification decision. If not, the maximum softmax output is oppressed by the resulting almost equal distributed class probabilities.
Both ODIN and the baseline have the advantage that no further changes to the network architecture are required.

\subsection{Learning Confidence Approach}
In comparison to the mentioned methods above, the confidence learning approach presented in~\cite{DeVries2018} changes the underlying network architecture by adding an additional confidence branch. The branch enables the network to output a degree of confidence $\gamma$ for a given input instead of just declaring an input sample as in- or out-of-distribution.
To be able to estimate the confidence inside this branch, the training procedure is changed as follows:
The classification output $o$ is interpolated with the one-hot encoded ground truth $y$, 
\begin{align}
    o' = \gamma\cdot o + (1-\gamma)\cdot y
\end{align}
where the degree of interpolation is the confidence of the network for the given input. In order to prevent the network from always state a low confidence and therefore get a low classification loss, a weighted confidence loss 
\begin{align}
    L_\gamma = -\lambda \cdot log(\gamma)
\end{align}
\vspace{-2pt}
is added to the classification loss. 
The weight $\lambda$ of the confidence loss is defined by a budget parameter $\beta$ and is adjusted whenever the weights are updated: If the confidence loss is greater than $\beta$, $\lambda$ is increased and the system is more punished for low confidences. Otherwise $\lambda$ is decreased and the system is getting less punished as a result of having a high confidence.

\vspace{-2pt}
\subsection{Metric Learning-based Approach}
\begin{figure}[!htb]
\centering
\scalebox{0.90}{
  \input{images/metriclearningnetwork.tikz}
}
\caption{Metric Learning-based approach to detect in- and out-of-distribution samples. The base network is extended by a Metric Learning layer ($f(x)$) as well as a confidence layer. The classification layer contains a residual block and is activated with a softmax.}
\label{fig:metriclearningnetwork}
\end{figure}
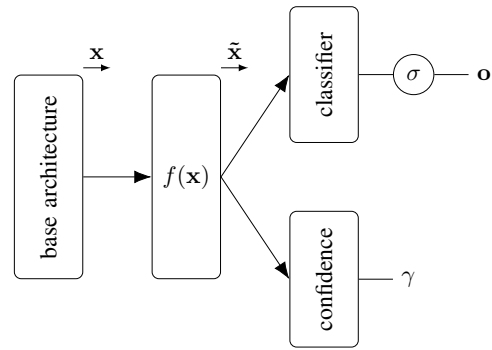

Like the confidence learning method, the approach based on Metric Learning changes the underlying network. 
More precisely, a Metric Learning layer ($f(x)$ in \autoref{fig:metriclearningnetwork}) is inserted between the base network and the classification layer.
Additionally a branch for learning the confidence, by approximating either the \textit{density} or \textit{entropy} of the learned manifold is appended.
The training is divided into two phases. First the Metric Learning layer is trained with the contrastive loss~\cite{Hadsell2006}. Afterwards the classification and confidence branches are trained on the resulting embeddings. The classification branch is trained straight forward by propagating the embeddings through a residual layer followed by a softmax activation. In contrast, the confidence approximation is a bit trickier.

The density as well as the entropy approximation use both the local neighborhood
\begin{align}
    n_\set{B}(x) = \{y \mid d(\tilde{x}, \tilde{y}) < m\}, & \; &  x, y \in \set{B}, \; m \in \mathbb{R}^{+}
\end{align}
\vspace{-2pt}
of an embedded sample $\tilde{x}$ in a batch $\set{B}$ to calculate the appropriate score. The neighborhood is given by all other samples in the batch where the (Euclidean) distance $d(.,.)$  to the corresponding embedding is lower than a predefined margin $m$.

\subsubsection{Density Approximation}
One of the simplest method on detecting how dense the area around a given sample $x$ is, is by calculate the local density 
\begin{align}
    \rho_\set{B}(x) = \frac{\left|n_\set{B}(x)\right|}{\left|\set{B}\right|}
\end{align}
\vspace{-2pt}
of the neighborhood, where a normalization factor is directly given by the batch size.
Indeed, using this calculation as the ground truth whilst the confidence training, the network learns to approximate the density but still lacks of the information about the pureness of the area. In an unclear decision region, the network should be able to give additional information, especially in terms of decision confidence.
The entropy approach tries to fix this issue.

\subsubsection{Entropy Approximation}
Instead of simply calculating the density, the entropy (\autoref{eq:entropy})
or Gini impurity (\autoref{eq:gini}) provide information about the purity of the local neighborhood.
\begin{align}
    H(Y) =& - \sum_{c \in \set{C}} p(c \mid Y) \cdot log(p(c \mid Y))
    \label{eq:entropy}
\\
    G(Y) =& 1 - \sum_{c \in \set{C}} p(c \mid Y)^2
    \label{eq:gini}
\end{align}
\vspace{-2pt}
where $\set{C}$ is the set of all available classes.
The approach is similar to the density approximation but requires a few tweaks in the ground truth calculation. 
Since the Gini impurity (the entropy) reaches its minimum (maximum) if all samples belong to the same class, a weighting needs to take care of empty neighborhoods. This refers to neighborhoods consisting only of the processed sample itself.
The weighting term for a sample $x$ is therefore given by
\begin{align}
    \omega_\mathcal{B}(x) = \frac{(\left|n_\set{B}(x)\right| - 1) \cdot \left|\set{B}\right|}{1 + \sum_{z \in \set{B}} \left|n_\set{B}(z)\right|}
\end{align}
and weights neighborhoods according to their size.
After applying the weighting term, the resulting loss for the Gini impurity and entropy approximation is
\begin{align}
    L_{H} = \frac{1}{\left| \mathcal{B} \right|} \cdot \sum_{x \in \mathcal{B}} \left| \gamma - \omega_\mathcal{B}(x) \cdot H(\mathcal{B})\right|
\end{align}
which is the mean of the absolute differences between the calculated ground truth values for the batch $\mathcal{B}$ and the networks confidence output $\gamma$.

\section{Evaluation}
\label{sec:eval}
The following section first describes the pipeline and introduce the novel semi-synthetic dataset. Afterwards the evaluation metrics are explained. Finally, the results are presented in a quantitative way.

\subsection{Pipeline}
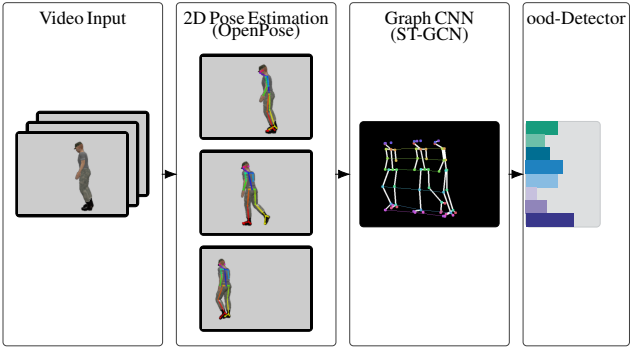
\begin{figure}[htb]
    \centering
    \scalebox{0.35}{
        \input{images/pipeline/pipeline.tikz}
    }
    \caption{Basic pipeline for all experiments. First, the video input is divided into single frames. Those frames are then analyzed by an 2D pose estimator (e.g. OpenPose). The resulting skeleton sequences are then propagated through a graph CNN (e.g. ST-GCN) and finally analyzed by an OoD-detector.}
    \label{fig:pipeline}
\end{figure}

As already mentioned, this work does not focus on the application of OoD-detector methods in the image domain. Instead, the applicability to graph-based data is examined. As an example, the problem of action recognition is analyzed where the input data is given in form of a sequence of graph skeleton data.
\autoref{fig:pipeline} shows the basic pipeline, which is similar to the one presented in~\cite{Yan2018}.
Given a video input, single frames are extracted and analyzed by a 2D pose estimator (e.g. OpenPose~\cite{Cao2018}). The resulting sequence of skeleton data is then propagated through a graph CNN (e.g. ST-GCN~\cite{Yan2018}) resulting in a regularized high-level representation of the input data. 
Based on this extracted high-level representation, the OoD-detectors are examined and the classification is done.

\subsection{Semi-synthetic Dataset}
\label{sec:dataset}
\begin{figure*}[tb]
    \centering
    \subfloat[]{
        \includegraphics[width=.3\linewidth]{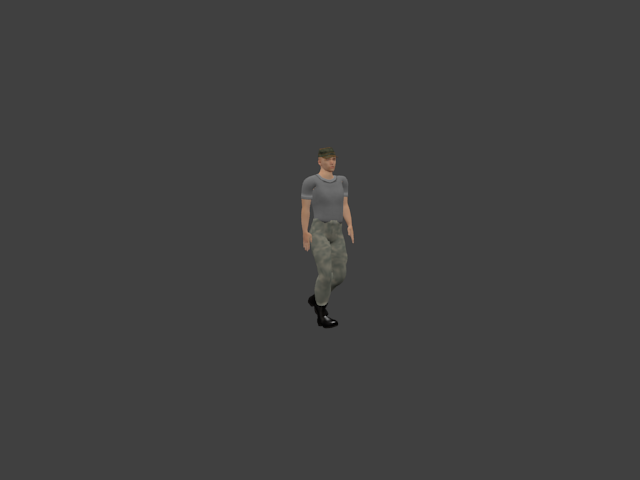}
    }
    \subfloat[]{
        \includegraphics[width=.3\linewidth]{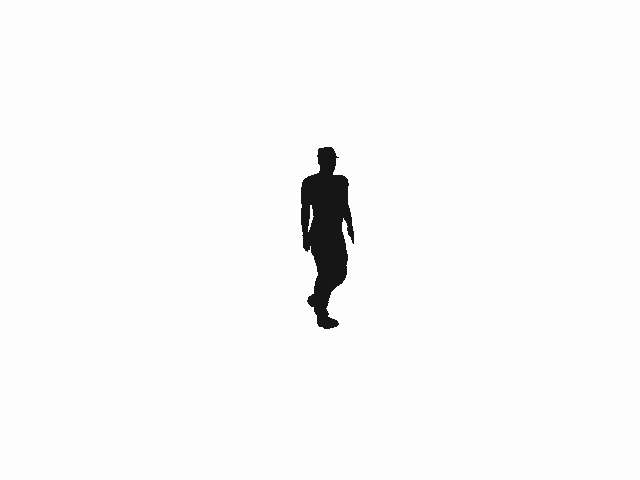}
    }
    \subfloat[]{
        \includegraphics[width=.3\linewidth]{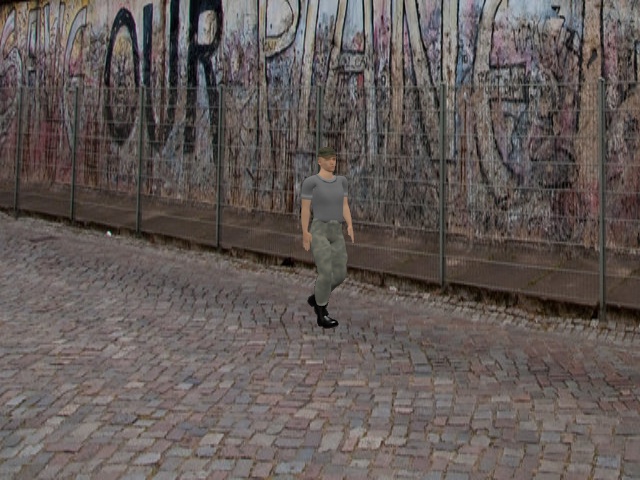}
    }
    \caption{Example image of a sequence of the semi-synthetic dataset: The same image as (a) RGB image, (b) depth image and (c) RGB image with a modified background.}
    \label{fig:datasetexampleimages}
\end{figure*}

\begin{figure}[tb]
    \centering
    \includegraphics[width=\linewidth]{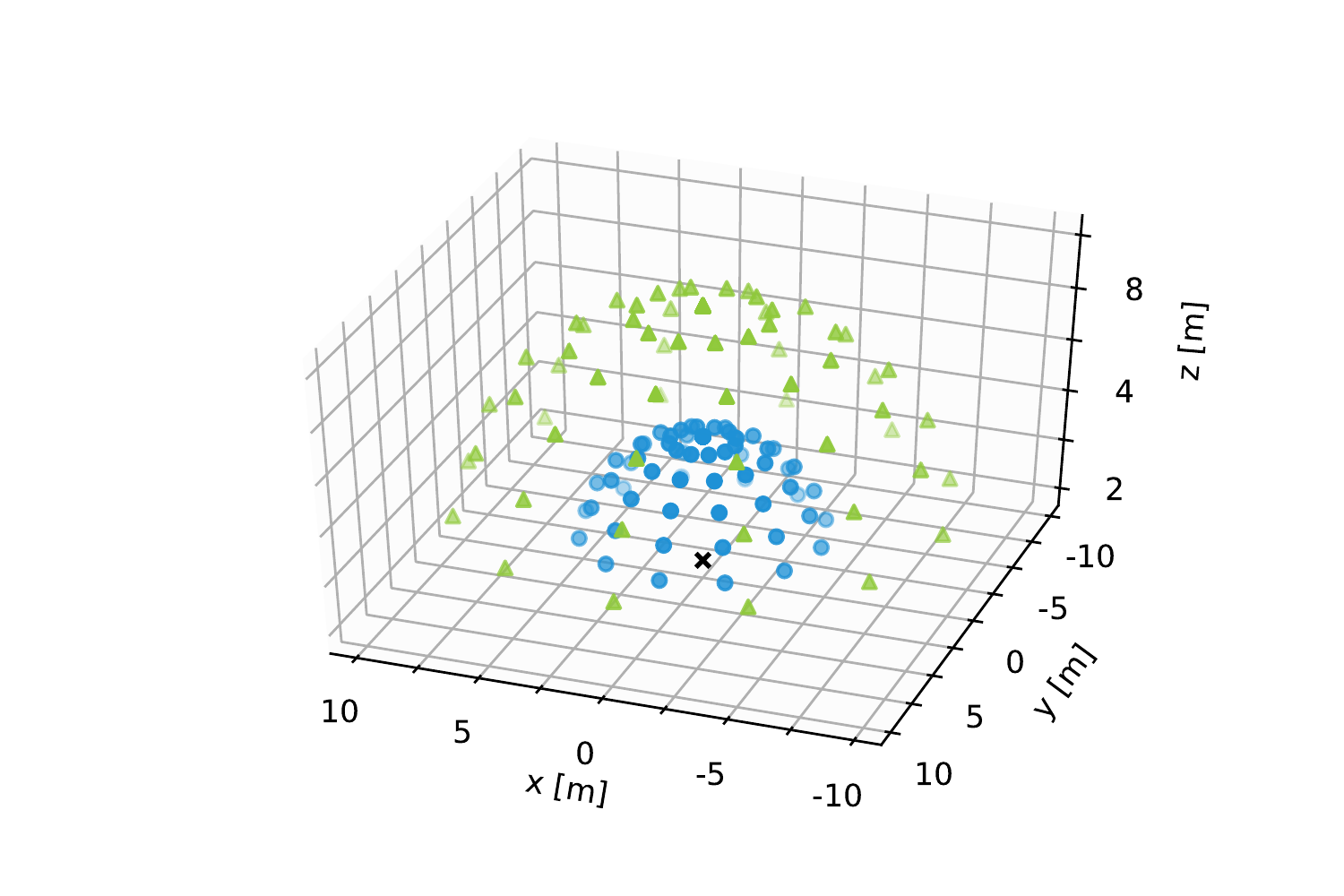}
    \caption{Camera positions during the rendering. All cameras (blue dots and green triangles) are directed towards the scene center (black cross). The blue cameras have a distance of 5m to the scene center while the green ones have a distance of 10m.}
    \label{fig:camerapositions}
\end{figure}

To obtain reproducible results, a novel semi-synthetic dataset is used. The dataset provides a controllable environment and is based on skeleton data of the CMU Graphics Lab Motion Capture Database~\cite{mocapCMU}. This skeleton data is used to animate a human 3D model~\cite{Makehumancommunity}. The resulting sequences are rendered with Blender~\cite{BlenderOnlineCommunity2019} from 144 different camera settings (\autoref{fig:camerapositions}). This can be seen as data augmentation and enables a scale and viewpoint invariance of the network~\cite{Bayer2020}. Each rendered RGB image has a resolution of $640\stimes480$px, depth data in form of a corresponding $640\stimes480$px 16bit-grayscale image (\autoref{fig:datasetexampleimages}) and a 18 node ground truth skeleton (\autoref{fig:datasetgraphbasedata}).
Currently, there are 32 different classes of actions in 109 sequences.\\
To verify the results and be able to check on interdataset OoD-samples, the NTU RGB+D~\cite{Shahroudy2016} dataset is used additionally.
It contains 60 action classes, presented in RGB videos with a resolution of $1920\stimes1080$px each, recorded from three different viewpoints. Compared to the short basic actions of the novel synthetic dataset, the NTU-RGB+D dataset contains more complex actions in which several persons may be involved.

\subsection{Metrics}
There are four established metrics used for the comparison of the different approaches~\cite{Hendrycks2016, Liang2017, DeVries2018}. The first one is the false positive rate (FPR) when the true positive rate (TPR) is fixed at $95\%$ (FPR 95). The second one is the detection error at the same fixed true positive rate. The area under the receiver-operator characteristic (AUROC) and the area under the precision-recall curve (AUPR) are the last two.

\subsubsection*{FPR 95}
The FPR 95 measures the false positive rate when the true positive rate is fixed at $0.95\%$.

\subsubsection*{Detection Error}
The detection error measures the misclassification probability when a a fixed true positive rate at $95\%$ is given. More precisely, the error is given by:
\begin{align}
    P_e = \sfrac{1}{2} (1-TPR + FPR)
\end{align}

\subsubsection*{AUROC}
The receiver-operator characteristic compares the true positive rate of a classifier with the corresponding false positive rate. The area under the receiver-operator characteristic is a threshold independent metric, measuring the overall performance of a classifier.

\subsubsection*{AUPR}
Another threshold independent metric is the area under the precision-recall curve. Unlike the AUROC, the AUPR is more sensitive to imbalanced datasets which is a desirable feature when examining OoD-detectors. Since the inlier and outliers can both be handled as positives in the AUPR calculation, a AUPR-IN and AUPR-OUT score is given respectively.

\subsection{Experimental Setup}
This work distinguishes between an interclass and interdataset OoD-detection. 
For the interclass case, only the semi-synthetic dataset is taken into account. For each of the 32 different classes and each detector, a network is trained. In each training, a single class represents the OoD-class and is excluded from the training whilst the other 31 classes are in-distribution classes and included in the training. 
Therefore this problem can be labelled as novelty detection.
For the interdataset case, the trained networks from the interclass OoD-detection were investigated on how good they distinguish between the 31 semi-synthetic (inlier) classes and the NTU-RGB+D plus the selected semi-synthetic (outlier) classes.

The data is split according to a stratified cross-validation into a test- and training set in a ratio of 1:4. As data augmentation, the skeleton graphs are modified by the following pipeline:
First a random start and end point of a sequence is defined, guaranteeing a sequence length of 20 frames. Then a Gaussian noise $(\mu=0, \sigma=0.005)$ is added to the node values. After this, there is a $50\%$ chance that nodes will be set to zero (kind of a dropout) and a $50\%$ chance that a vertical and horizontal mirroring is also applied. The noise as well as the application of dropout and the mirroring is fix for a whole sequence.

The human skeleton graphs are extracted by OpenPose using the default settings. 
ST-GCN with the spatial configuration partitioning strategy has been chosen for the analysis of the resulting graphs~\cite{Yan2018}.
The ST-GCN networks are all initialized with random values. Adam was used as optimizer with its default parameters except for the initial learning rate.

For the evaluation, the procedure described in~\cite{Hendrycks2016} has been followed. First the test set is separated into correctly and incorrectly classified examples. From the two resulting groups, the AUROC and AUPR scores are calculated. Afterwards, the confidence threshold is estimated in such a way, that the true positive rate of the correctly classified examples drops to $95\%$. 
Based on this threshold, the FPR and detection error is calculated.
Since there are 32 different classes and therefore 32 trained networks for a given method, the results are averaged and the corresponding standard deviation is given.

\subsubsection{Baseline and ODIN}
The baseline method and ODIN do not require a modification of the existing network, so they can easily be examined without retraining. However, in order to be able to compare these straightforward approaches without interference from the smarter architectures, they are analyzed with the base architecture.
The networks are trained over 100 epochs with a batch size of 512. The initial learning rate is set to 0.001 and shrinks all 40 epochs by a factor of 10.

\subsubsection{Learning Confidence}
Like the previous networks, the learning confidence system is also trained over 100 epochs. The batch size is set to 1024 and the initial learning rate is 0.001 and shrinks all 30 epochs by a factor of 10. The budget parameter is set to 0.3.

\subsubsection{Metric Learning}
The training of the Metric Learning approach is divided into three parts. First the Metric Learning layer is trained to get a good representation of the data in the manifold. Based on the embedding, the classifier and OoD-detector are trained, while the weights of the Metric Learning layer are being held fixed.

The Metric Learning layer is trained over 200 epochs with a batch size of 512. The learning rate also starts at 0.001 and shrinks every 80 epochs by a factor of 10. The layer maps the input onto a 256 dimension output.
The classification layer and the confidence layer are then both separately trained over 50 epochs with an initial learning rate of 0.0001 and a reduction every 20 epochs by a factor 10.

\subsection{Results}
\label{sec:results}
In the following, the results are presented in quantitative terms. To give a hint where the value $0.95$ resides, each of the following plots contains a dotted red line. The temperature parameter is displayed logarithmically on the x-axis. \autoref{tbl:intraclass-cmp} and \autoref{tbl:intradataset-cmp} provide results of the intraclass and intradataset evaluation. The row of the ODIN method in both tables contains the values of the ODIN parameterization with the lowest valid FPR score.

\subsubsection{Baseline and ODIN}
The results for the baseline as well as ODIN are shown in \autoref{fig:odin-results}. The parameter $T=1$ equals the baseline. In the intraclass OoD-detection, the FPR reaches its minimum at a temperature parameter of $T=1.6$. For temperature parameter values above 100, the required TPR of $95\%$ cannot be guaranteed and are therefore not taken into account. Compared to the results in~\cite{Liang2017}, the curve has an unusual course for a rising temperature parameter. The curve of the intradataset, on the other hand, shows a similar course as the results in~\cite{Liang2017}.

\subsubsection{Learning Confidence}
\begin{figure}[t]
     \centering
     \includegraphics[width=\linewidth]{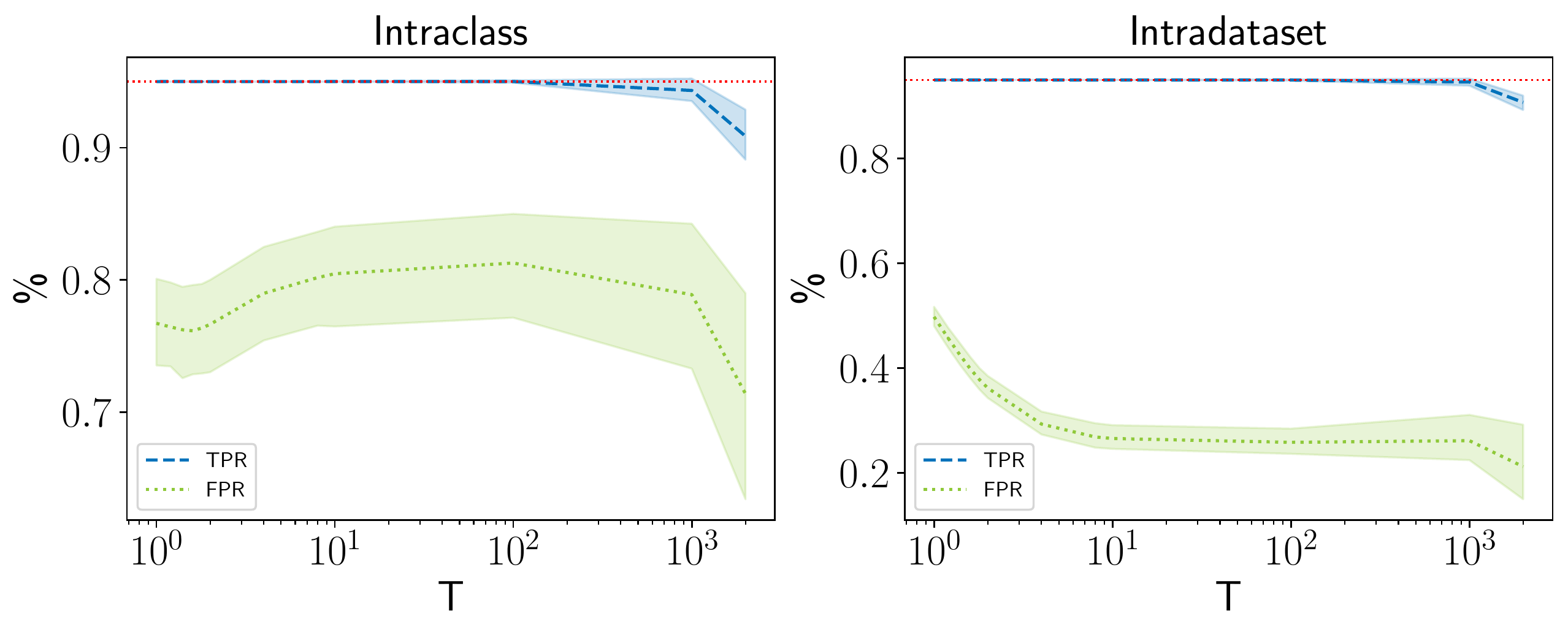}
     \caption{ODIN Results: TPR and FPR for different temperature parameters. At $T=100$, the TPR is no longer fixed at $95\%$, which leads to a better FPR but also allows more errors.}
     \label{fig:odin-results}
 \end{figure}

\begin{figure}[t]
     \centering
     \includegraphics[width=\linewidth]{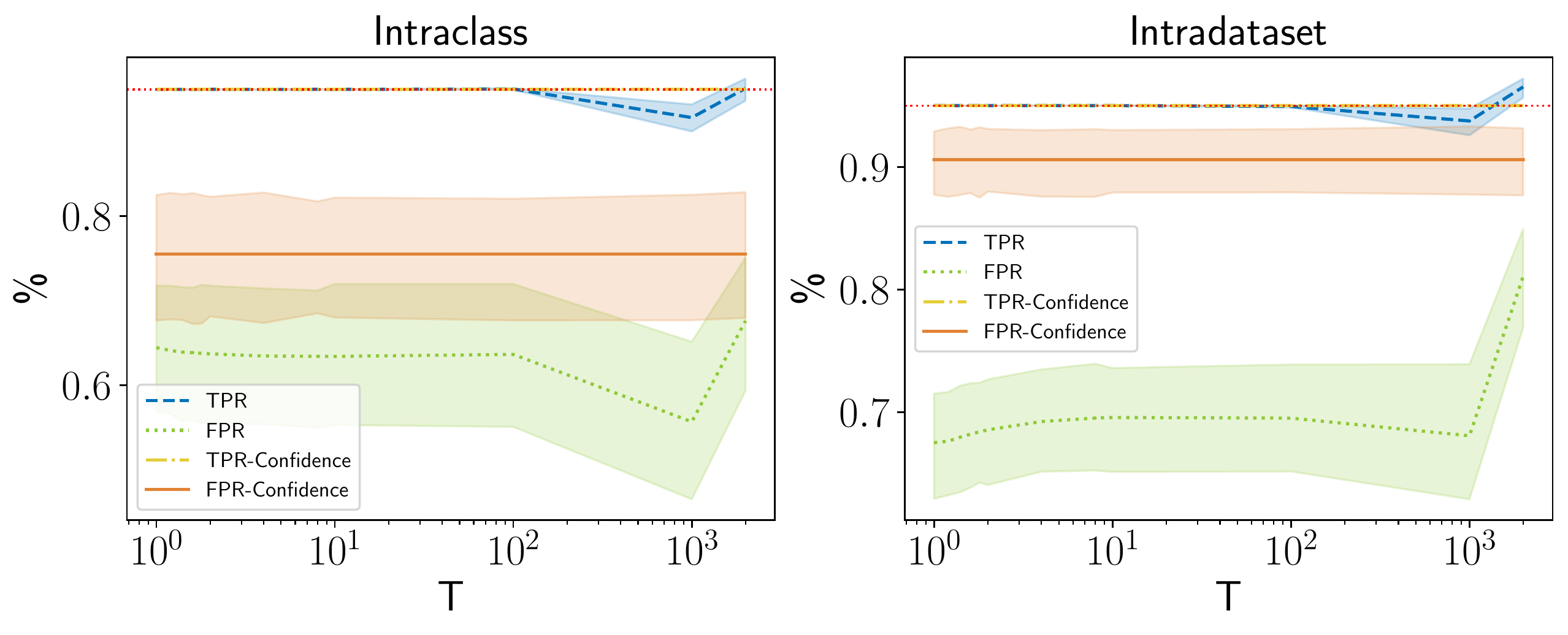}
     \caption{Learning Confidence Results: It is noticeable, that the FPR as well as the TPR increase for $T=2000$, which is a strange behavior compared to the other plots.}
     \label{fig:confidence-results}
 \end{figure}

\autoref{fig:confidence-results} shows the results of the confidence learning approach. The intradataset ODIN curve vary heavily from the ones, depicted in \autoref{fig:odin-results}. It should also be noted, that for the intradataset case the method performs significantly worse than ODIN, even if ODIN operates on the modified network architecture. Another remarkable problem is, that the average accuracy not included in the plots has dropped from $0.74\pm0.03$ (base architecture) down to $0.62\pm0.09$ (learning confidence).

\subsubsection{Metric Learning}
\begin{figure}[t]
     \centering
     \includegraphics[width=\linewidth]{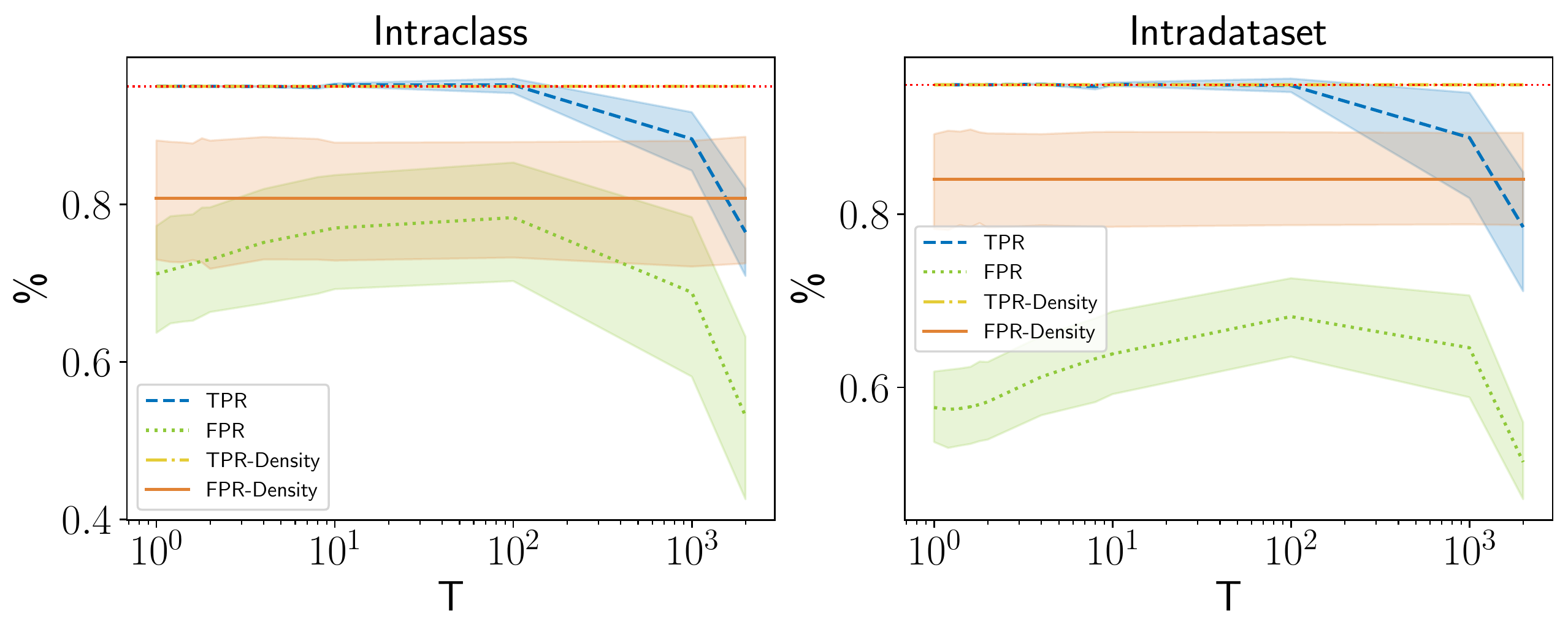}
     \caption{Metric Learning Results: Density approximation. The intraclass and intradataset FPR values for the density approach differ primarily in the standard deviation.}
     \label{fig:metric-results-density}
 \end{figure}

\begin{figure}[t]
     \centering
     \includegraphics[width=\linewidth]{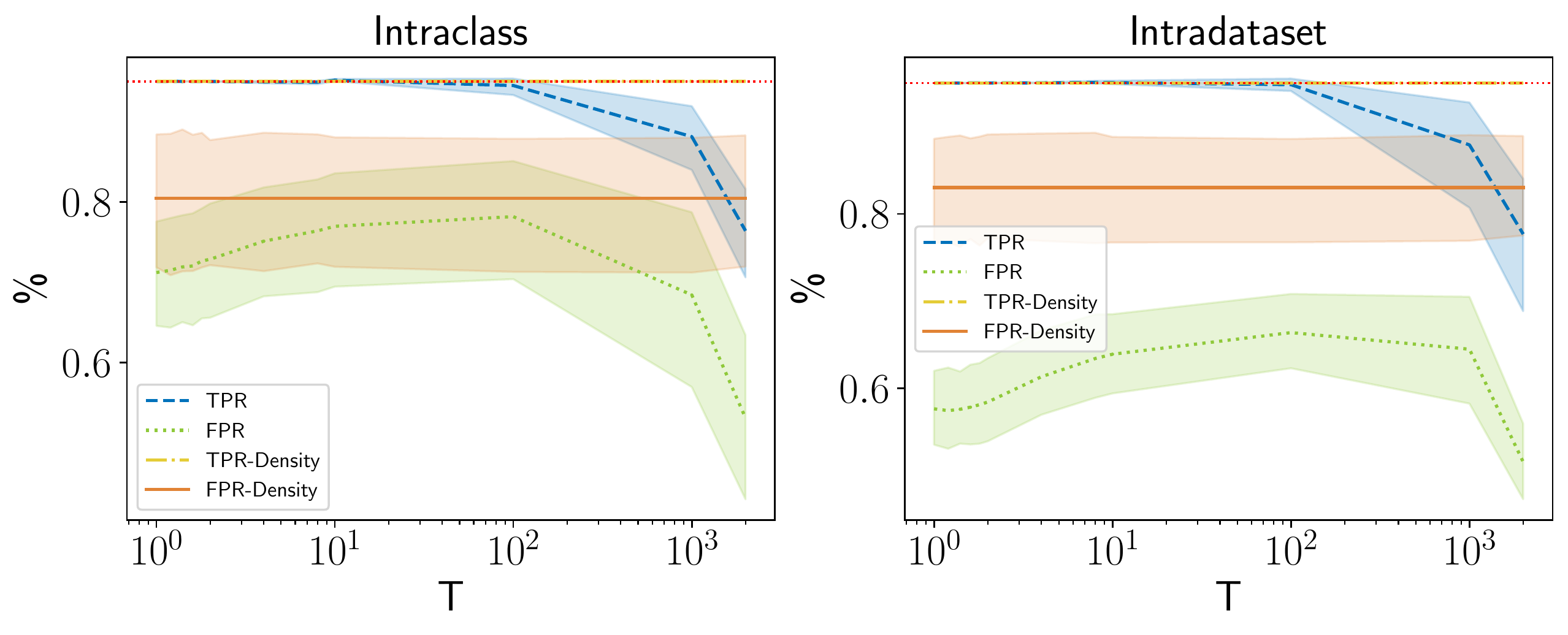}
     \caption{Metric Learning Results: Entropy approximation. As the density approximation, the FPR values of the entropy approach differ primarily in the standard deviation.}
     \label{fig:metric-results-entropy}
 \end{figure}

The density approximating as well as the entropy approximating Metric Learning approaches were investigated and result in similar plots  (\autoref{fig:metric-results-density}, \autoref{fig:metric-results-entropy}) for the intraclass and intradataset comparison. Therefore the following analysis can be applied to both of them.\\
The ODIN intraclass FPR curve has some similarities with the curve of the base architecture. Nonetheless, the TPR curve drops at $T=100$ much heavier than in the base architecture. In comparison to the learning confidence approach and in terms of the intraclass task, both Metric Learning approaches perform worse than the learning confidence or ODIN. In terms of the intradataset task, they perform slightly better than the learning confidence approach but also have a higher variance. The average accuracies of the density and entropy classifiers are $0.648\pm0.12$ and $0.647\pm0.12$ and therefore also slightly better than the learning confidence ones ($0.62\pm0.09$).

%


\begin{table*}[!hb]
    \renewcommand{\arraystretch}{1.3}
    \centering
    \caption{Intraclass comparison: Training and test on the same dataset. A single class is excluded from the training\newline and serves as OoD-class in the test phase.}
    \label{tbl:intraclass-cmp}
    \begin{tabular}{|l|c|c|c|c|c|c|}
    \hline
    Method & FPR $\downarrow$ & AUROC $\downarrow$ & AUPR-IN $\uparrow$ & AUPR-OUT $\uparrow$ & ERR $\downarrow$ & ACC $\uparrow$ \\
    \hline
    Baseline & $0.77\pm0.10$ & $0.77\pm0.10$ & $0.94\pm0.08$ & $0.30\pm0.21$ & $0.41\pm0.05$ & $0.74\pm0.03$\\
    ODIN & $0.76\pm0.10$ & $\mathbf{0.77\pm0.10}$ & $\mathbf{0.94\pm0.07}$ & $0.30\pm0.21$ & $0.41\pm0.05$ & $\mathbf{0.74\pm0.03}$\\
    Confidence & $\mathbf{0.76\pm0.22}$ & $0.76\pm0.15$ & $0.94\pm0.08$ & $\mathbf{0.35\pm0.27}$ & $\mathbf{0.40\pm0.11}$ & $0.62\pm0.09$\\
    Density & $0.81\pm0.23$ & $0.65\pm0.23$ & $0.89\pm0.15$ & $0.29\pm0.27$ & $0.43\pm0.12$ & $0.65\pm0.12$\\
    Gini & $0.80\pm0.25$ & $0.67\pm0.23$ & $0.89\pm0.13$ & $0.29\pm0.27$ & $0.43\pm0.13$ & $0.65\pm0.12$\\
    \hline
    \end{tabular}
\end{table*}

\begin{table*}[!hb]
    \renewcommand{\arraystretch}{1.3}
    \centering
    \caption{Intradataset comparison: Training on the semi-synthetic dataset, test on the NTU-RGBD+D dataset.}
    \label{tbl:intradataset-cmp}
    \begin{tabular}{|l|c|c|c|c|c|c|}
    \hline
    Method & FPR $\downarrow$ & AUROC $\downarrow$ & AUPR-IN $\uparrow$ & AUPR-OUT $\uparrow$ & ERR $\downarrow$ & ACC $\uparrow$ \\
    \hline
    Baseline & $0.50\pm0.06$ & $0.88\pm0.02$ & $0.74\pm0.05$ & $0.95\pm0.01$ & $0.27\pm0.03$ & $0.74\pm0.03$\\
    ODIN & $\mathbf{0.26\pm0.07}$ & $0.94\pm0.02$ & $\mathbf{0.85\pm0.08}$ & $\mathbf{0.98\pm0.00}$ & $\mathbf{0.15\pm0.04}$ & $\mathbf{0.74\pm0.03}$\\
    Confidence & $0.91\pm0.08$ & $0.53\pm0.04$ & $0.16\pm0.03$ & $0.88\pm0.03$ & $0.48\pm0.04$ & $0.62\pm0.09$\\
    Density & $0.84\pm0.16$ & $\mathbf{0.49\pm0.18}$ & $0.25\pm0.16$ & $0.82\pm0.10$ & $0.45\pm0.08$ & $0.65\pm0.12$\\
    Gini & $0.83\pm0.18$ & $0.52\pm0.18$ & $0.26\pm0.17$ & $0.84\pm0.10$ & $0.44\pm0.09$ & $0.65\pm0.12$\\
    \hline
    \end{tabular}
\end{table*}

\section{Conclusion}
\label{sec:conclusion}
As the evaluation shows, OoD-methods can successfully be applied to graph-based data, but their behavior is different as on image-based data.
Experiments showed, that the OoD-detector method ODIN outperforms the more sophisticated learning confidence and the metric learning based method. Despite the modification of the network architecture made by these two methods, ODIN is superior when used with the modified architecture. As in \autoref{tbl:intraclass-cmp} and \autoref{tbl:intradataset-cmp} is shown, the modified network architectures even have a negative impact on the classification accuracy (ACC). This is of particular interest as the learning confidence method outperforms ODIN in the original paper in nearly every case. Another interesting observation is, that the learning confidence method performs better in the intraclass task than the intradataset task.

In this paper we have shown with our novel semi-synthetic dataset, that applying ODIN on graph-based data is currently the best OoD-method.

Our presented metric learning based method embeds the high-level features into a manifold and learns to estimate the density or entropy of the local neighborhood of an embedded sample. Since it is crucial to find a good embedding, other embedding methods than the contrastive loss could be investigated in future work.

\section*{Acknowledgment}
This work was developed in Fraunhofer Cluster of Excellence ``Cognitive Internet Technologies''.




%
\bibliographystyle{IEEEtran}
\bibliography{bibliography}

\end{document}

%% file: images/metriclearningnetwork.tikz
\begin{tikzpicture}[]
  \newcommand{\fully}[2]{
  \draw[-{Latex[scale=1.5]}] (#1.east) to (#2.west);
  }

  \node (base) [draw, rounded corners=1mm, rectangle, minimum width=1cm, minimum height=3cm, anchor=west] at (0,0) {\rotatebox{90}{base architecture}};
  \node (metric) [draw, rounded corners=1mm, rectangle, minimum width=1cm, minimum height=3cm, anchor=west] at ($(base.east)+(1,0)$)  {$f(\vec{x})$};
  \node (cls) [draw, rounded corners=1mm, rectangle, minimum width=1cm, minimum height=2cm, anchor=south west] at ($(metric.east)+(1,0.5)$) {\rotatebox{90}{classifier}};
  \node (cnf) [draw, rounded corners=1mm, rectangle, minimum width=1cm, minimum height=2cm, anchor=north west] at ($(metric.east)+(1,-0.5)$) {\rotatebox{90}{confidence}};

  \node [anchor=west] (softmax) [draw, circle, anchor=west] at  ($(cls.east)+(0.5,0)$) {$\sigma$};
  \node [anchor=west] (o) at  ($(softmax.east)+(0.5,0)$) {$\mathbf{o}$};
  \node [anchor=west] (c) at ($(cnf.east)+(0.5,0)$) {$\gamma$};

  \fully{base}{metric}
  \fully{metric}{cls}
  \fully{metric}{cnf}

  \draw (cls) -- (softmax) -- (o);
  \draw (cnf) -- (c); 
      
  \draw [->, >=latex]  ($(base.north east)+(0,+0.1)$) -- node [anchor=south]{$\vec{x}$} ($(base.north east)+(0.4,+0.1)$);
  \draw [->, >=latex]  ($(metric.north east)+(0,+0.1)$) -- node [anchor=south]{$\vec{\tilde{x}}$} ($(metric.north east)+(0.4,+0.1)$);
\end{tikzpicture}

%% file: images/pipeline/pipeline.tikz.tex
\begin{tikzpicture}
      \newcommand{\connect}[2]{
        \draw[-{Latex[scale=2]}] (#1.east) to (#2.west);
      }
    \definecolor{fgreen}{RGB}{23,156,125}
    \definecolor{fgreen2}{RGB}{109,191,169}
    \definecolor{fblue}{RGB}{31,130,192}
    \definecolor{fblue2}{RGB}{136,188,226}
    \definecolor{fpurple}{RGB}{57,55,139}
    \definecolor{fpurple2}{RGB}{144,133,186}
    \definecolor{fpurple3}{RGB}{199,193,222}
    \definecolor{fgray2}{RGB}{199,202,204}
    \definecolor{flblue}{RGB}{0,110,146}

    \begin{scope}[local bounding box=videoinput, outer xsep=4mm]
      \node (videobox) [draw=darkgray, rounded corners, minimum width=6cm, minimum height=13cm] at (0,0) {};
      \node [anchor=north, align=center, outer ysep=2mm, text width=6cm] at (videobox.north) {\huge{Video Input}};
      \node [draw, rounded corners, black, fill](base_0) at (-0.4,0,-2)
      {\includegraphics[width=4cm]{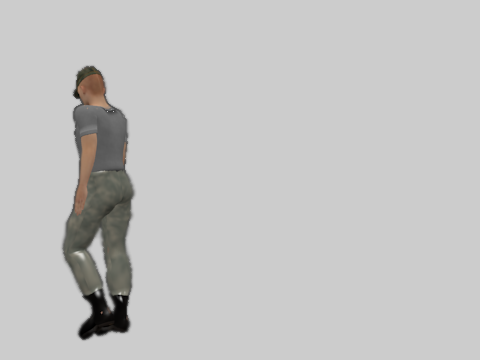}};
      \node [draw, rounded corners, black, fill](base_1) at (-0.4,0,-1) {\includegraphics[width=4cm]{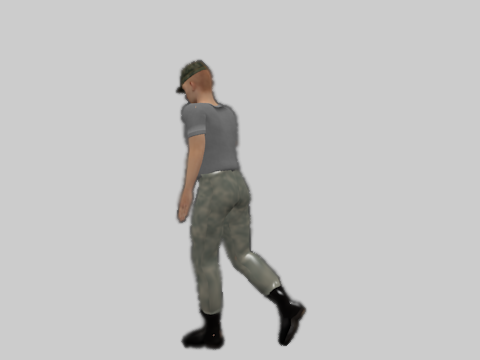}};
      \node [draw, rounded corners, black, fill](base_2) at (-0.4,0,0)
      {\includegraphics[width=4cm]{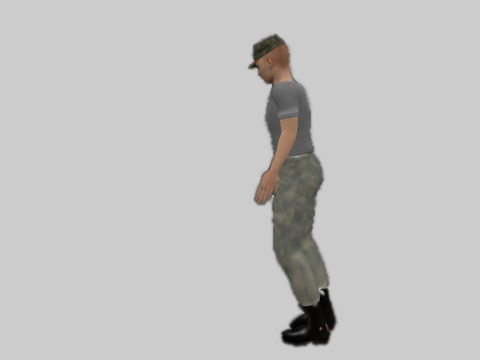}};
    \end{scope}
    \begin{scope}[shift=(videoinput.east), anchor=west, outer xsep=4mm, local bounding box=openpose]
      \node (openposebox) [draw=darkgray, rounded corners, minimum width=6cm, minimum height=13cm] at (0,0) {};
      \node (openposetext) [anchor=north, outer ysep=2mm, text width=6cm, align=center] at (openposebox.north) {\huge 2D Pose Estimation (OpenPose)};
      \node [draw, rounded corners, black, fill, anchor=north, outer sep=2mm] (pose_0) at (openposetext.south) {\includegraphics[width=4cm]{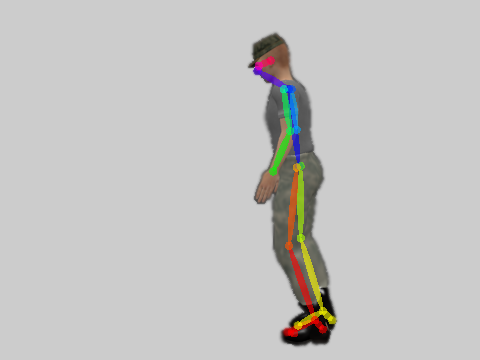}};
      \node [draw, rounded corners, black, fill, anchor=north, outer sep=2mm] (pose_1) at (pose_0.south) {\includegraphics[width=4cm]{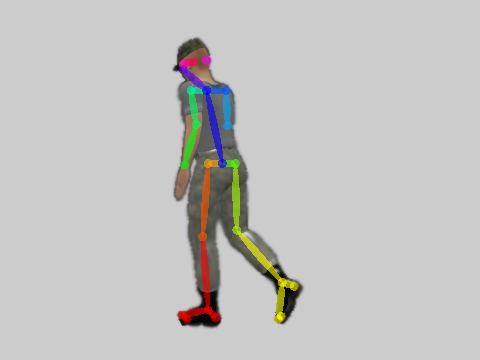}};
      \node [draw, rounded corners, black, fill, anchor=north, outer sep=2mm] (pose_2) at (pose_1.south) {\includegraphics[width=4cm]{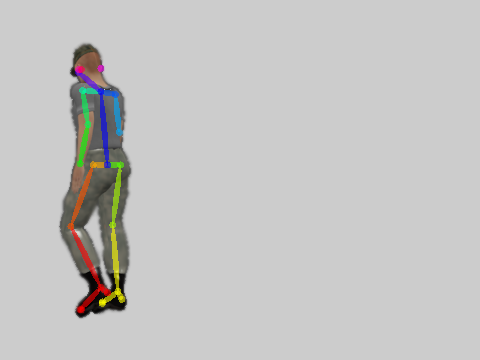}};
    \end{scope}
    \begin{scope}[shift=(openpose.east), anchor=west, outer xsep=4mm, local bounding box=stgcn]
      \node (stgcnbox) [draw=darkgray, rounded corners, minimum width=6cm, minimum height=13cm] at (0,0) {};
      \node (stgcntext) [anchor=north, outer ysep=2mm, text width=6cm, align=center] at (stgcnbox.north) {\huge Graph CNN (ST-GCN)};
      \node [draw, rounded corners, black, fill, anchor=center] (skeleton) at (stgcnbox.center) {\includegraphics[width=5cm]{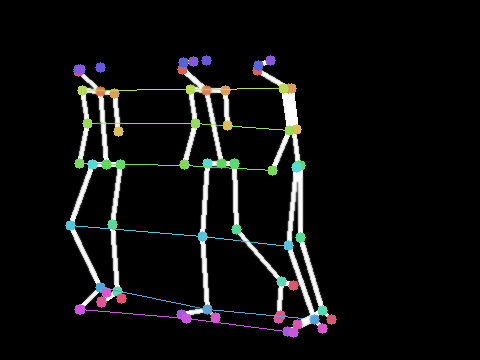}};
    \end{scope}
    \begin{scope}[shift=(stgcn.east), anchor=west, outer xsep=4mm, local bounding box=classifier]
      \node (classifierbox) [draw=darkgray, rounded corners, minimum width=4cm, minimum height=13cm] at (0,0) {};
      \node (classifiertext) [anchor=north, outer ysep=2mm] at (classifierbox.north) {\huge{ood-Detector}};
      \begin{scope}[outer xsep=5mm]
        \node [anchor=west, rounded corners, draw, fgray2, fill=fgray2!60, minimum width=2.8cm, minimum height=4cm] (cls) at (0,0) {};
        \node (c1) [fill=fgreen, minimum width=12mm, minimum height=5mm, anchor=north west, outer ysep=0] at (cls.north west) {} ;
        \node (c2) [fill=fgreen2, minimum width=7mm, minimum height=5mm, anchor=north west, outer ysep=0] at (c1.south west) {} ;
        \node (c3) [fill=flblue, minimum width=9mm, minimum height=5mm, anchor=north west, outer ysep=0] at (c2.south west) {} ;
        \node (c4) [fill=fblue, minimum width=14mm, minimum height=5mm, anchor=north west, outer ysep=0] at (c3.south west) {} ;
        \node (c5) [fill=fblue2, minimum width=12mm, minimum height=5mm, anchor=north west, outer ysep=0] at (c4.south west) {} ;
        \node (c6) [fill=fpurple3, minimum width=4mm, minimum height=5mm, anchor=north west, outer ysep=0] at (c5.south west) {} ;
        \node (c7) [fill=fpurple2, minimum width=8mm, minimum height=5mm, anchor=north west, outer ysep=0] at (c6.south west) {} ;
        \node (c8) [fill=fpurple, minimum width=18mm, minimum height=5mm, anchor=north west, outer ysep=0] at (c7.south west) {} ;
      \end{scope}
    \end{scope}
    
    \draw[-{Latex[scale=2.5]}] ([xshift=-1.5mm]videoinput.east) -- ([xshift=0.4cm] videoinput.east);
    \draw[-{Latex[scale=2.5]}] ([xshift=-1.5mm]openpose.east) -- ([xshift=0.4cm] openpose.east);
    \draw[-{Latex[scale=2.5]}] ([xshift=-1.5mm]stgcn.east) -- ([xshift=0.4cm] stgcn.east);
\end{tikzpicture}